\documentclass[final]{cvpr}

\usepackage{times}
\usepackage{epsfig}
\usepackage{graphicx}
\usepackage{amsmath}
\usepackage{amssymb}
\usepackage{siunitx}
\usepackage{grffile}

\usepackage{gensymb}

\usepackage[pagebackref=true,breaklinks=true,colorlinks,bookmarks=false]{hyperref}

\def\cvprPaperID{10511}
\def\confYear{CVPR 2021}

\begin{document}

\title{Nutrition5k: Towards Automatic Nutritional Understanding of Generic Food}

\author{Quin Thames \and Arjun Karpur \and Wade Norris$^\ast$ \and Fangting Xia \and Liviu Panait \and Tobias Weyand \and Jack Sim \and
\\
Google Research, USA \\
{\tt\small \{quinbob,arjunkarpur,sukixia,liviu,weyand,jacksim\}@google.com} \\
Perception Labs\thanks{Work done while Wade Norris was at Google.} \\
{\tt\small wnorris@perceptionlabs.ai}

}
\maketitle

\begin{abstract}

Understanding the nutritional content of food from visual data is a challenging computer vision problem, with the potential to have a positive and widespread impact on public health. Studies in this area are limited to existing datasets in the field that lack sufficient diversity or labels required for training models with nutritional understanding capability. We introduce Nutrition5k, a novel dataset of 5k diverse, real world food dishes with corresponding video streams, depth images, component weights, and high accuracy nutritional content annotation. We demonstrate the potential of this dataset by training a computer vision algorithm capable of predicting the caloric and macronutrient values of a complex, real world dish at an accuracy that outperforms professional nutritionists. Further we present a baseline for incorporating depth sensor data to improve nutrition predictions. We release Nutrition5k in the hope that it will accelerate innovation in the space of nutritional understanding. The dataset is available at \small{\url{https://github.com/google-research-datasets/Nutrition5k}}.

\end{abstract}

\section{Introduction}
\label{introduction}

\begin{figure}[h]
\includegraphics[width=1.0\columnwidth]{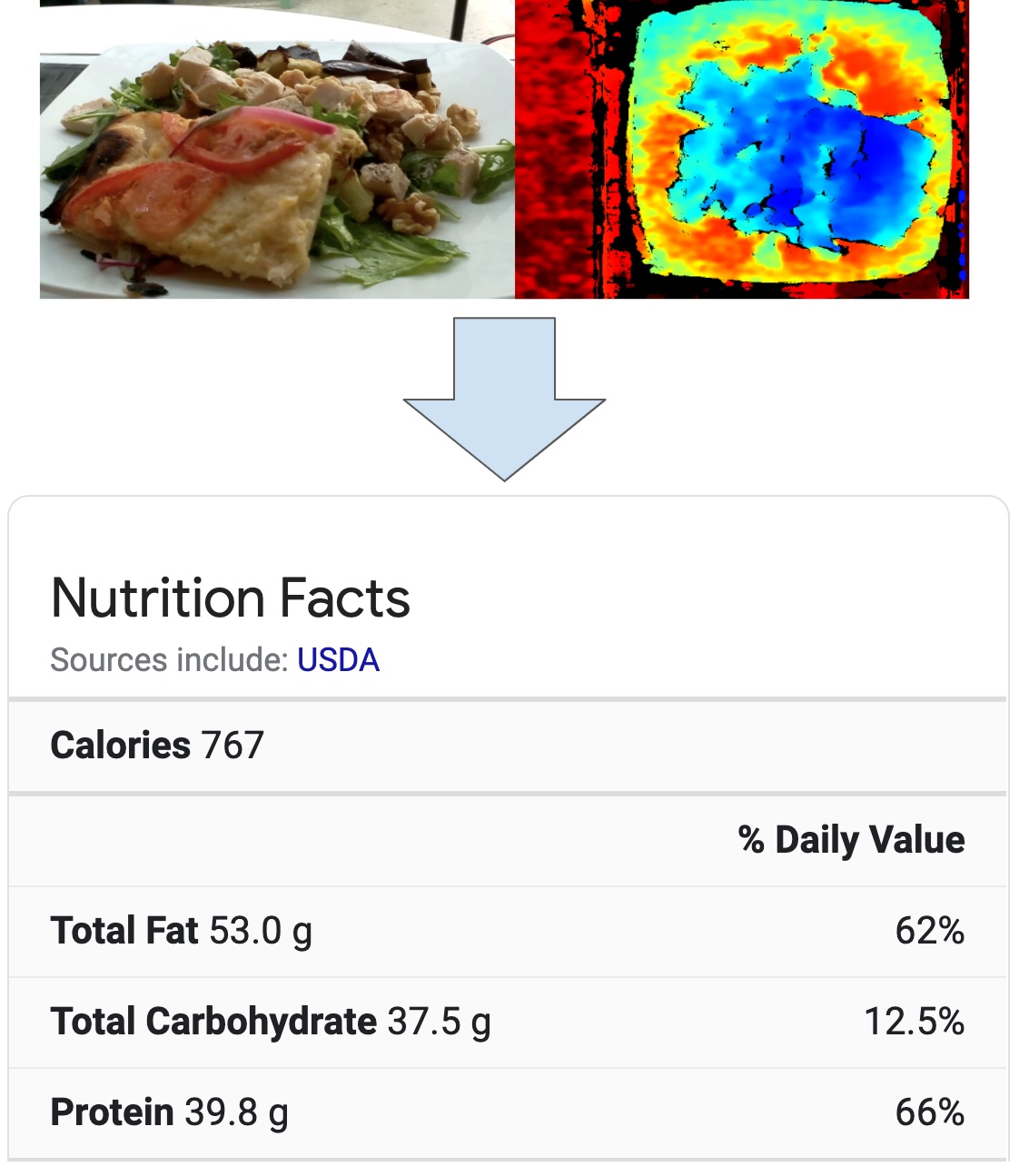}
\caption{ Example representation of data contained in \textit{Nutrition5k}. }
\label{figure-overview}
\end{figure}

The nutritional composition of a person's diet is inextricably linked to health, happiness, and longevity. Making it easier to understand and track the nutritional breakdown of the food we eat enables us to make better dietary choices and potentially live longer and healthier lives. Despite the considerable impact that what we eat has on us, the tools for trying to understand food are extremely cumbersome and limited.

Currently, the main approach for an individual who wants to record the nutritional content of their food intake is to utilize a tracking app such as MyFitnessPal. These types of apps enable a user to set intake goals for total calories, as well as for specific macronutrients (carbohydrate, protein, and fat), and provide an interface to log and aggregate meals to see if these goals are met. When doing this type of tracking, each ingredient or dish consumed must be individually logged along with the exact portion size. These logs can then be automatically converted to nutritional content using average per portion values saved in a database. Many apps require a scale to individually weigh and log each ingredient the user eats, but this can be a tedious and time consuming process. Short of using a scale, users can also attempt to perform visual portion size estimation themselves, but this can be highly error-prone step ~\cite{lee2012comparison, schap2011adolescents}. MyFitnessPal boasted over 19M monthly unique active users in early 2018 \cite{HealthFitnessApps} despite these difficulties in data entry. The ability to streamline food logging with a camera would not only make this process easier for current active users, but it may also unlock another large demographic of potential users.

As the impact of nutrition on our health grows more apparent, more progress has been made in making this data more accessible. Many chain restaurants now post the nutritional content of menu items online and in stores. A variety of promising techniques have leveraged this publicly available source of data to show the potential for classification models to help with food logging. Such approaches demonstrate a dramatic reduction in effort when logging meals by enabling a user to take a photo of their dish rather than manually entering nutritional content. However, this requires that the dish belong to a set of known menu items with presumably fixed portion sizes and nutritional content. Frequently, though, our meals are not on a list of known dishes and current capabilities in this space are extremely limited.

We use the term \textit{generic food} for dishes that are not necessarily from a pre-determined set with known nutritional content. Understanding the content of generic food requires the ability to estimate portion sizes in addition to recognizing the ingredients present, making it a significantly more difficult problem than classification alone.

One of the largest challenges to furthering progress in this space is data collection. For many computer vision problems, researchers are able to leverage the abundance of data readily available on the internet to train their models. Unfortunately for the nutritional understanding space, data on the internet is sparse and often inaccurate. Images from many sources are often plated and shot with an artistic intention to increase appeal rather than represent realism. The few large scale and diverse datasets, such as Recipe1M \cite{marin2018learning}, are mined from recipe websites. While these contain valuable dish level, ingredient level, and preparation attribute annotations, they almost always lack annotations for the portion sizes shown in the photos. Without accurate portion size annotations, learning to predict the nutritional content from the associated images is difficult and error prone. A potential solution for building a nutrition dataset could be to send web-mined images to human annotators and have them estimate the portion size. However, our findings show that this annotation task is extremely difficult, even for nutrition experts, and produces highly inaccurate labels.

This paper explores an alternative approach to dataset construction: incrementally weighing, scanning, and logging each item as it’s added to a plate in real world cafeterias immediately before consumption. A variety of weight and imagery sensors are used for the scanning process and the ingredient breakdown is logged via the item recipe, enabling the calculation of near exact nutrition annotations. The end result is Nutrition5k, a dataset of five thousand unique, real world, generic food dishes and their associated video captures, depth images, component weights, and high accuracy nutritional information. We demonstrate the utility of this dataset by training deep CNNs to predict nutritional content from a single RGB image, achieving an accuracy that even surpasses a trained nutritionist's ability to do so visually. Additionally, we make use of depth sensor data as an additional signal to greatly improve our portion size predictions and nutrition regression accuracy.

\section{Related Work}
\label{related-work}

Many prior works~\cite{bolanos2016simultaneous, mezgec2017nutrinet, im2calories, singla2016food} follow the approach of classifying a dish into one of a known set of menu items, where each instance of an item is assumed to have approximately the same nutritional breakdown. Such an approach is scalable to some extent (as many as 2,500 unique dishes are classified in \cite{im2calories}), but is limited to classifying an entire dish image as a single food item. Bolanos et al. \cite{bolanos2016simultaneous} extended this to support images with multiple dishes present at once by using a two stage system that first detects individual dishes, and then classifies those regions.

Another set of techniques utilize the abundance of web data for recipes to help understand images of dishes~\cite{chen2016deep, min2018you, salvador2017learning, ruede2020multitask}. These approaches leverage web labels to train embedding models that retrieve similar recipes, or use the attribute labels for classification. However, they make no attempt at portion estimation and therefore aim to predict relative nutritional information but cannot compute the absolute nutritional content from the given dish image.

Approaches such as \cite{fang2015single, fang2016comparison} explicitly tackle the portion estimation problem by fitting geometric shapes to detected food items in order to predict volume. These approaches provide accurate results but focus on datasets with limited variation. Other works employ progress in dense labeling techniques to perform portion annotation via segmentation. Myers et al.~\cite{im2calories} combine a segmentation network with a depth prediction network and convert voxels to food volume estimates, but their evaluation of this method is limited to plastic food replicas used for training nutritionists. Ciocca et al. \cite{cioccaJBHI} introduce a dataset with 1,027 tray images and segmentation masks for 73 food categories but with no portion annotations.

The authors of~\cite{im2calories} claim that once food volume has been retrieved, nutritional density datasets can be used to calculate mass, calorie, and macronutrients. Ando et al.~\cite{ando2019depthcaloriecam} build on~\cite{im2calories} by using an on-device smartphone depth sensor to estimate volume from overhead imagery and learning food item density to predict mass, but are ultimately limited in their food density data and only show successful experiments for three food items. Similarly, Gao et al.~\cite{gao2019musefood} use on-device sensors to estimate food volume but make no attempts to determine food density/mass.

Liang et al. \cite{liang2017portion} present a dataset with 2,978 images of 160 unique pieces of food and use a FasterRCNN network to predict the mass. These are mostly whole foods such as an entire apple or tomato, one per plate. Many of these approaches also utilize a fudicial marker for scale; for example \cite{liang2017portion} uses a Yuan coin and \cite{fang2015single} uses a checkered square.

\section{Dataset}

\begin{table}[t]
\begin{center}
\begin{tabular}{|l|cccc|}
\hline
Name & Unique & Mixed & Depth & Public\\
\hline
MenuMatch\cite{beijbom2015menu} & 41 & Y & N  & Y \\ %
ECUSTFD\cite{liang2017portion} & 160 & N & N & Y \\ %
Fang et al.\cite{fang2015single} & 3 & Y & N & N \\ %
Ando et al.\cite{fang2015single} & 3 & Y & Y & Y \\ 
\textit{Nutrition5k} [Ours] & 5,066 & Y & Y & Y \\ %
\hline
\end{tabular}
\end{center}
\caption{Summary of nutrition datasets (i.e. contain portion and calorie annotations). Here we define unique as the number of dishes with a unique combination of ingredients so that it has a different underlying nutritional breakdown (e.g. different ingredients or different portion sizes thereof). Mixed refers to multiple ingredients per dish.
}
\label{table-datasets}
\end{table}

\subsection{Goals}

Nutrition5k is intended to be a diverse dataset of food dishes with in-depth, high accuracy annotations, and scale sufficient to train a high capacity neural network. The hope is that this will spur innovative research into new approaches to automatic nutritional understanding in the generic food space and provide a benchmark for future evaluation. The following were important motivations when collecting the dataset:

\begin{itemize}
    \item \textbf{Diversity} For an approach to generalize to the culinary diversity seen in the real world, the training dataset must cover a wide variety of ingredients, portion sizes, and dish complexities.
    \item \textbf{Realistic} By capturing photographs of real food, in actual campus cafeterias, the data simulates real world conditions where logging meals would occur.
    \item \textbf{Challenging} As in real world dishes, images in Nutrition5k contain ambiguities and challenges that may not be possible to fully overcome (e.g. ingredients may be partially or fully occluded by others).
    \item \textbf{High accuracy} To surpass human ability in visual nutrition estimation, it is important to gather data that gives a more accurate understanding of the contents of a plate (i.e. depth images and component weighing) than visual estimation alone.
\end{itemize}

\noindent\textbf{Non-goals} The intent of this dataset is to focus on the more challenging generic food space, and not to replace approaches where a dish can be classified into a set of known menu items with known portion sizes, as is done in some of the aforementioned related works \cite{bolanos2016simultaneous, mezgec2017nutrinet, im2calories, singla2016food}. Retrieval and classification style approaches, when possible, will inherently produce higher accuracy results than those intended for generic food, and as such we see the challenges as distinct.

\begin{figure}[h]
\includegraphics[width=\columnwidth]{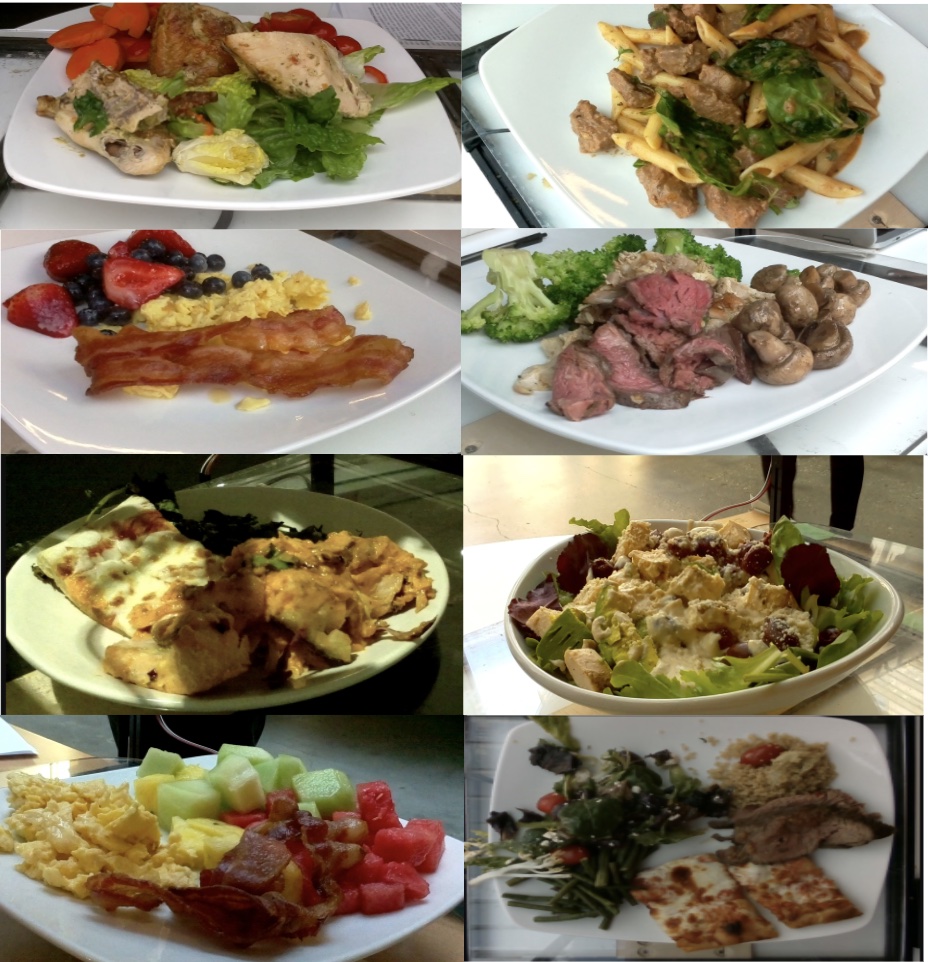}
\caption{ RGB image examples from \textit{Nutrition5k}.}
\label{figure-example-images}
\end{figure}

\begin{figure}[h]
\includegraphics[width=\columnwidth]{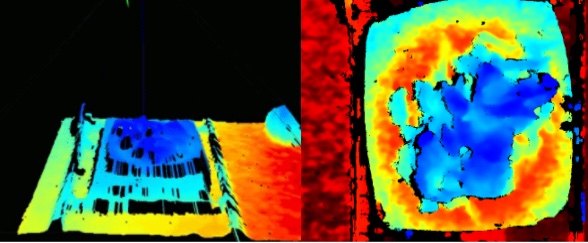}
\caption{ Depth image examples from \textit{Nutrition5k}.}
\label{figure-depth-example-images}
\end{figure}

\subsection{Scale and Splits}

Nutrition5k contains 20k short videos generated from roughly 5000 unique dishes constructed from more than 250 different ingredients. Each dish has a full breakdown of ingredient labels, their quantities, and their macronutrient information computed using the USDA Food and Nutrient Database~\cite{food2010usda}. Furthermore, 3.5k out of the 5k dishes also include overhead RGB-D images captured from an Intel RealSense camera. As far as we are aware, Nutrition5k is the largest nutrition dataset with portion annotations.

The data is divided into training and testing subsets. We partition 10\% of the dishes into the test set, and leave the remainder as train. We call these corresponding sets Nutri-Train and Nutri-Test.

\begin{figure*}[h]
\begin{center}
\includegraphics[width=2.0\columnwidth]{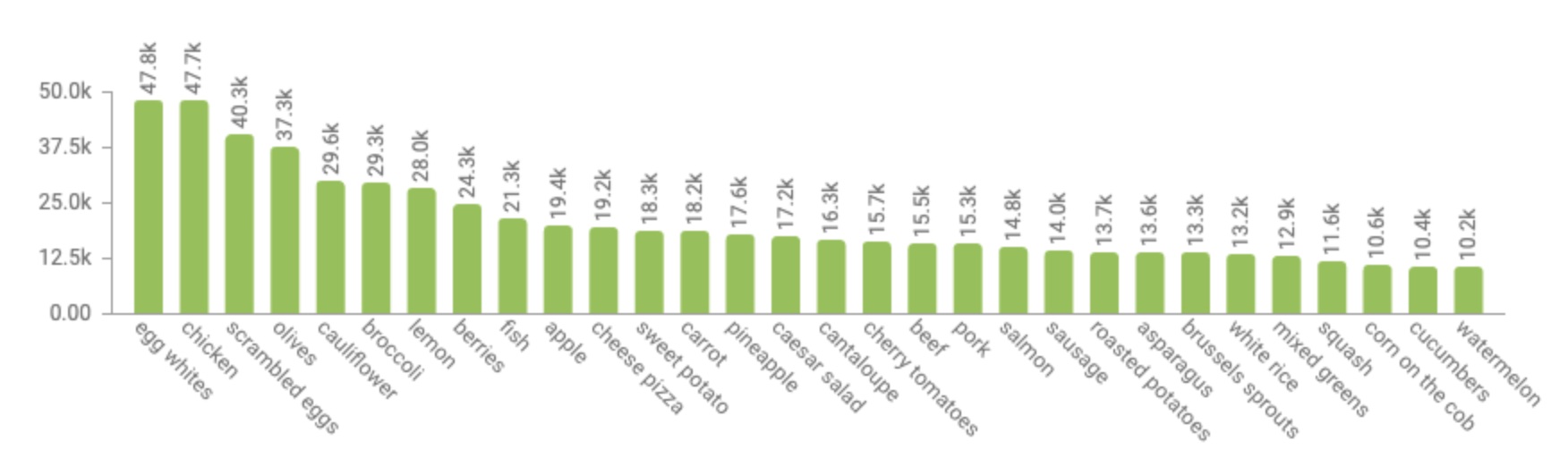}. %
\caption{ Top 30 ingredients by mass in \textit{Nutrition5k}. }
\label{figure-label-counts}
\end{center}
\end{figure*}

\subsection{Data Distribution}
The dishes in Nutrition5k vary drastically in portion sizes and dish complexity, with dishes ranging from just a few calories to over 1000 calories and from a single ingredient to up to 35 with an average of 5.7 ingredients per plate. Table~\ref{table-metric-stats} shows the variety in nutritional breakdown of the dishes used. Figure~\ref{figure-label-counts} shows the 30 most common ingredients in Nutrition5k by mass. Sorting ingredients by mass instead of frequency gives a more intuitive look at the common ingredients used, as the most frequently occurring ingredients are basics such as salt, pepper, olive oil, and vinegar.

\begin{table}[t]
\begin{center}
\begin{tabular}{|l|c|c|c|}
\hline
& Mean & Standard & Average Deviation\\
& & Deviation & from Mean\\
\hline
Calorie & 255 & 220 & 136\\
Total Mass(g) & 215 & 161 & 114\\
Fat(g) & 12.7 & 13.5 & 6.93\\
Carbs(g) & 19.4 & 21.6 & 10.3\\
Protein(g) & 18.0 & 20.0 & 10.7\\
\hline
\end{tabular}
\end{center}
\caption{Mean, standard deviation, and average deviation from mean for each metric in \textit{Nutrition5k}.}
\label{table-metric-stats}
\end{table}

\subsection{Supervision Labels}

Our \textit{Nutrition5k} dataset can be represented as ${D}=\{{I}_i, {Y}_i\}_{i=1}^N$, in which
${I}_i$ is the image, ${Y}_i$ is the supervision labels, and $N$ is the number of examples. 
The supervision labels ${Y}_i=(y_i^w, {Y}_i^m, y_i^{cal})$ 
consist of three types: total weight label $y_i^w$, macronutrient labels 
${Y}_i^m$, and a calorie label $y_i^{cal}$. Note that all of the supervision labels are a function of the weight (in grams) of each ingredient ${K}_l$ where ${l}$ is an index running over all ingredients as follows:
\begin{itemize}
    \item $y_i^w$ represents the total weight of each dish. 
    \item ${Y}_i^m$ is a vector of the weights of each macronutrient, i.e. Let ${M}=\{carb, fat, protein\}$, then  $\forall j \in {M}$, $y_{ij}^m = F_{macro}({K}_l, j)$ where $F_{macro}$ is a function that calculates the amount of each macronutrient.
    \item $y_i^{cal} = F_{calorie}({K}_l)$ where $F_{calorie}$ is a function which calculates the total number of calories from per ingredient weights.
\end{itemize}

\subsection{Collection Procedure}
\begin{figure}[h]   
\includegraphics[width=\linewidth]{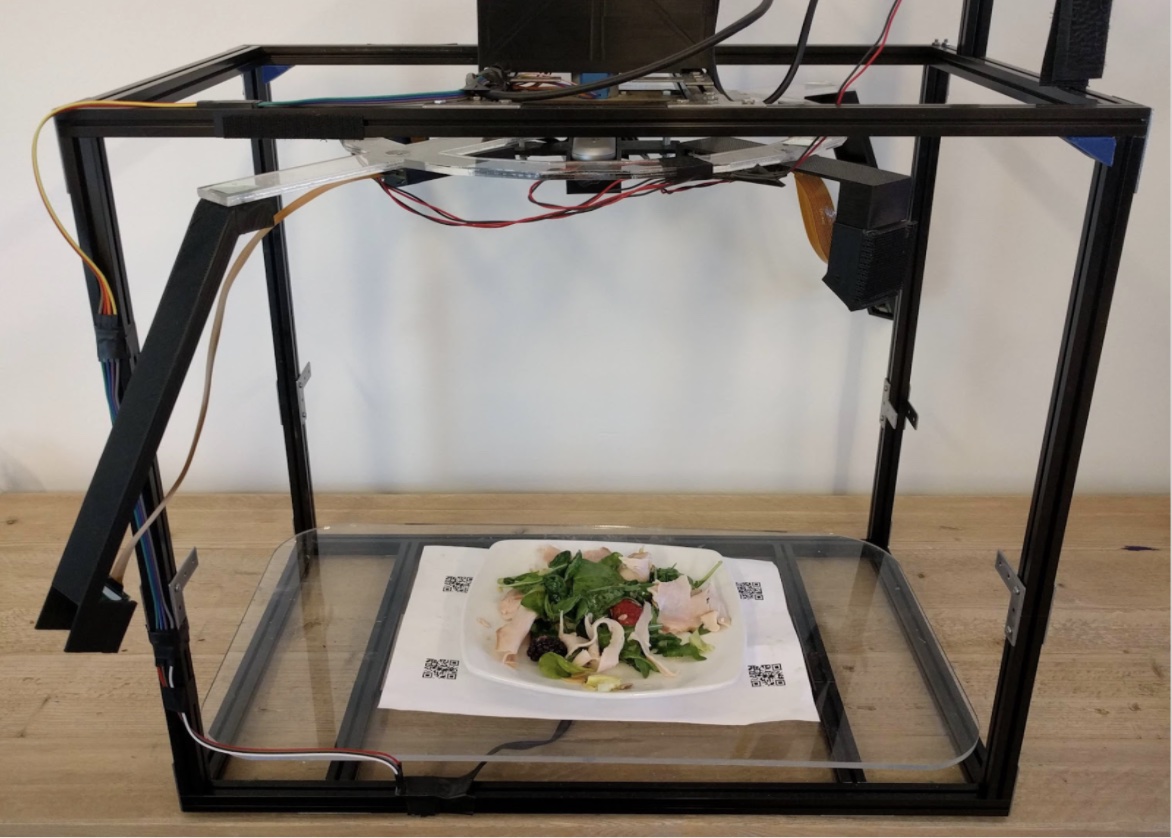}
\caption{ Robotic device for rapid data capture. }
\label{figure-capture-device}
\end{figure}

To capture and record data, we utilize a custom sensor array to weigh and scan each dish in Nutrition5k (see Figure~\ref{figure-capture-device}). Robotic automation was used to manipulate and trigger all of the sensors simultaneously to reduce the time required for each scan.

In curating complex dishes with accurate nutrition labels, items from the buffet style cafeterias were added one at a time to a plate or bowl, followed by a scan after each item was added. An item corresponds to either an individual ingredient or a prepared food with a known recipe, and thus known per gram ingredient breakdowns. Per gram nutrition content for each item was calculated using the per gram ingredient breakdown and the USDA Food and Nutrient Database. This value is then multiplied by the incremental weight measurement as the item is added to the plate to generate the ground truth annotation for nutrition content. As items were added, dishes grew from simple, single recipe plates to much more complex and challenging mixed recipe dishes. However, even a single item dish could be a quite complex recipe menu item.

Note that the data splits respect the incremental nature of scans, such that all dishes belonging to a single incremental scan will exist in either Nutri-Train or Nutri-Test. In this way, we ensure that the test and training sets do not have any overlapping images of the same plate.

For each scan, the following data was recorded: the recipe or ingredient of the item added, four RGB video recordings, an overhead RGB-D image, and an incremental weight measurement. The five cameras are oriented above and around the plate, with one pointing down from directly overhead and the remaining four from each side of the dishes. The four side-angle cameras all sweep 90 degrees simultaneously, capturing the full 360 degrees, and are alternated at approximately 30 degrees and 60 degrees down from the horizon. Each camera records in $1920 \times 1080$ resolution for approximately 8 seconds per sweep, and the depth image is generated by averaging over the 8 second period to remove noise and holes from the capture. We use an Intel RealSense D435 to collect overhead depth data, with depth units of $1\mathrm{e}{-4}$. A digital scale under the plate captures the incremental weight within +/- 1 gram of precision.

\subsection{Challenges}
While this dataset strives to achieve a high fidelity understanding of the nutritional content of dishes captured via per item mass and volume measurements, there are ways in which potential error could still accumulate. For example, picking out and removing specific ingredients from an established recipe could skew the nutrition content log, though such behavior was discouraged.

The complexity of collecting in depth annotations with specialized hardware limited our collection efforts for this dataset to a single real world cafeteria. While the cafeteria had a diverse and daily updating set of menu items from various cultural cuisines, the limited geographic nature of the collection inherently skews the food seen to mostly western style dishes. We leave adding scans from more geographically and culturally diverse cafes for future work.

There are also some potential biases introduced by leveraging robotic automation for efficient data collection. The relatively small set of viewing distances and the sensor support structures seen in the background could potentially be used by an algorithm to assist in portion size estimation. Our experiments show, however, that providing explicit depth information to models still proves to be a significant factor in portion estimation.

\section{Experiments}

This section details our experiments in developing computer vision models based on data collected in \textit{Nutrition5k}. We sample RGB frames and depth images from the dataset, and train models to predict the nutritional content of a dish. We present metrics to assess model performance based on prediction method, model architecture, and the subsequent impact each had on end-to-end nutritional understanding. We then compare our performance in portion estimation to estimates made by both expert (nutritionists) and novice annotators on a subset of the data.

\begin{figure*}
\includegraphics[width=\linewidth]{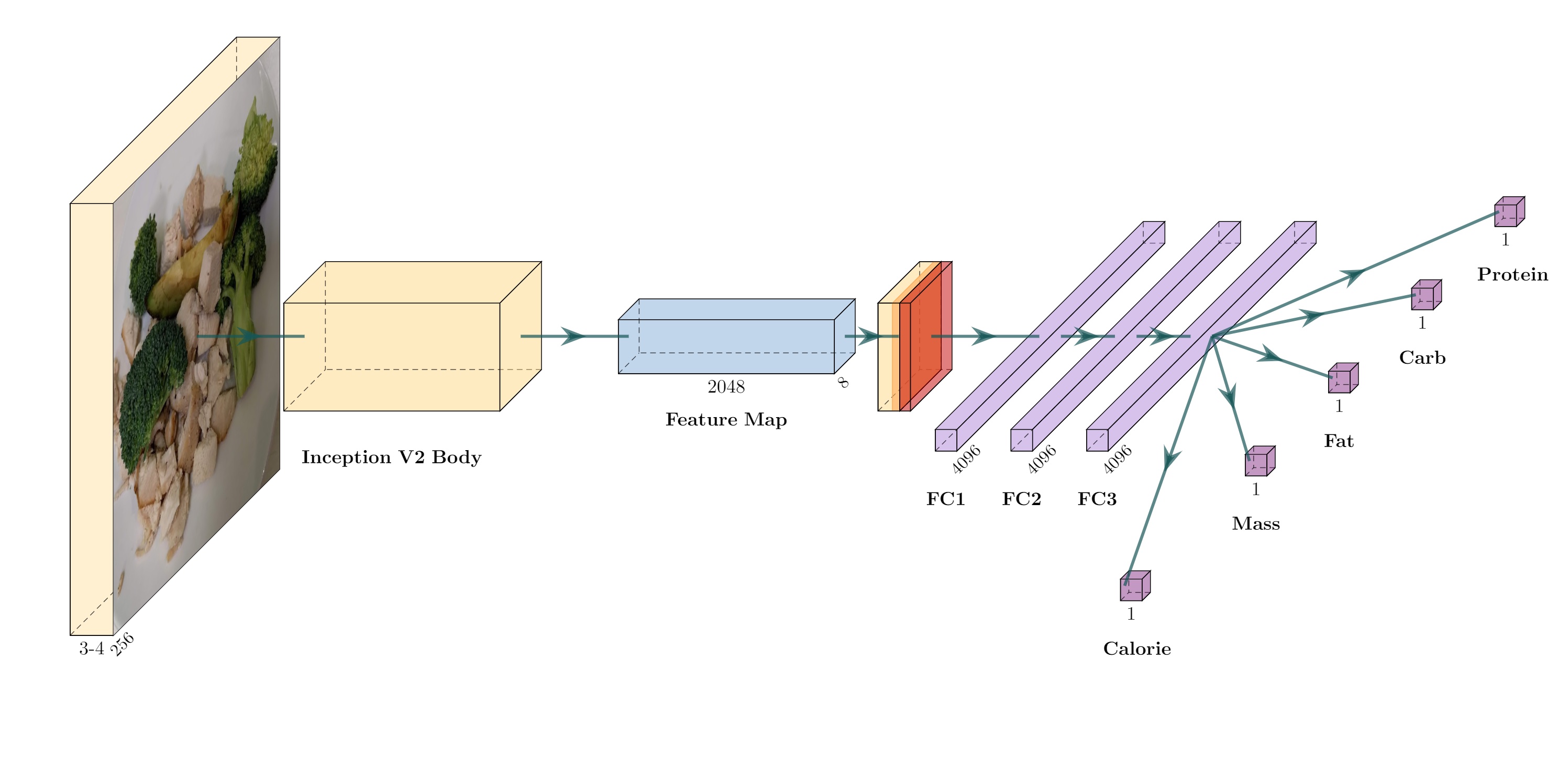}
\caption{ Overview of network architecture for our multi-task learning experiments. }
\label{figure-network-arch}
\end{figure*}

\subsection{Metrics}
Given the unique portion annotations and depth data available in \textit{Nutrition5k}, we conduct 3 novel experiments. First, we train a portion independent regression model where calories and macronutrients \textit{per gram of food} are predicted from a single RGB image. Next, we predict the precise nutritional content of a dish, portion (i.e. mass) included, from a single RGB image. Finally, we augment the RGB input to this direct prediction with depth data in an attempt to improve upon our 2D baseline. We measure the regression accuracy of calorie, total mass, and individual macronutrient mass using the mean absolute error (MAE), which is defined as:

\begin{equation}
MAE ={\frac{1}{N}\sum_{i=1}^{N}\left|\hat{y}_i-y_i\right|}
\end{equation}
where $\hat{y}_i$ is the predicted value for a given test image ${I}_i$ for each metric. Caloric values are measured in standard kilocalories units, while macronutrient and total masses are measured in grams. We present the MAE as a direct value in its respective units and as a percentage of the mean ground truth value.

\subsection{Experimental Setup}
\label{subsec:2D-image-models}

\noindent\textbf{Base architecture} Our architecture is based on an InceptionV2~\cite{szegedy2016rethinking} backbone encoder. The input resolution to the network is a 256x256 image, where images were downsized and center cropped in order to retain the most salient dish region. We optimize our network using the RMSProp algorithm, with an initial learning rate of $1\mathrm{e}{-4}$, momentum of $0.9$, decay of $0.9$, and epsilon of $1.0$. All models are pretrained using JFT-300M \cite{sun2017revisiting}.

\label{section-multitask}
\noindent\textbf{Multi-task learning} For each regression task (calorie, macronutrients, and optionally mass), we train a separate multi-task head. The multi-task learning architecture utilizes the mixed5c outputs from the InceptionV2 base and applies a [3, 3] average pooling kernel with stride 2 and valid padding. Two 4096-dimensional fully connected (FC) layers follow, and all tasks share weights up until this point. For each regression task, a final third and fourth FC layers follows (with dimension 4096 and 1, respectively) and the appropriate loss for the given task is used, as defined in Equation~\ref{eqn:multi_tasking_loss}. Figure~\ref{figure-network-arch} provides an overview of the network architecture.

We learn the weights ${W}$ for the multi-task network by minimizing the loss function $l_{multi}$ defined as:
{\small
\begin{align}
l_{multi}({D}|{W})=&\frac{1}{N}\sum_{i=1}^{N}[l_m({I}_i,{Y}_i^m|{W}) \nonumber\\
                            &+l_c({I}_i,y_i^{cal}|{W}) \nonumber\\ &+l_w({I}_i,y_i^{w}|{W})] \nonumber\\
l_m({I},{Y}^m|{W})=&\frac{1}{|{M}|}\sum_{j\in{M}}|\hat{y}_j^m - y_j^m| \nonumber\\
l_c({I},y^{cal}|{W})=&|\hat{y}^{cal} - y^{cal}| \nonumber\\ l_w({I},y^{w}|{W})=&|\hat{y}^{w} - y^{w}| \nonumber\\
\label{eqn:multi_tasking_loss}    
\end{align}
}
The overall loss $l_{multi}$ is a weighted combination of three sub-task loss functions: macronutrient regression loss $l_m$, calorie regression loss $l_c$, and total weight loss $l_w$. $l_m$, $l_c$, and $l_c$ use mean absolute error (MAE) as the regression loss. $\hat{y}_j^m$, $\hat{y}^{cal}$, and $\hat{y}^{w}$ are the predicted label values for the three sub-tasks while $y_j^m$, $y^{cal}$, and  $y^{w}$ are the ground-truth values.

\noindent\textbf{Experiment Data} The partitions between Nutri-Train and Nutri-Test remain constant for all experiments. For 2D only models, the rotating videos are sampled and every 5th frame is used for training. For the depth aware models, only the subset of dishes with RealSense RGB-D images are used.

\subsection{Nutrition Understanding from 2D Images}

\begin{table*}[t]
\begin{center}
\begin{tabular}{l|cccccc}
& Calorie MAE & Mass MAE & Fat MAE & Carb MAE & Protein MAE\\
\hline
Baseline & 150.8 / 60.2\% & 124.6 / 58.5\% & 8.2 / 67.6\% & 12.5 / 62.1\% & 10.5 / 62.1\% \\
\hline
2D Portion Independent Model & 24.1 / 9.5\% & - & 2.3 / 18.3\% & 2.7 / 13.9\% & 2.2 / 12.0\% \\
2D Direct Prediction & 70.6 / 26.1\% & 40.4 / 18.8\% & 5.0 / 34.2\% & 6.1 / 31.9\% & 5.5 / 29.5\% \\
\hline
Depth as 4th Channel & 47.6 / 18.8\% & 40.7 / 18.9\% & 2.27 / 18.1\% & 4.6 / 23.8\% & 3.7 / 20.9\% \\
Volume Scalar & 41.3 / 16.5\% & 29.4 / 13.7\% & 3.0 / 25.5\% & 4.5 / 22.0\% & 5.2 / 31.1\% \\
\end{tabular}
\end{center}
\caption{Mean absolute error (MAE) and mean absolute error as a percent of the respective mean for that field. Evaluations are run on test set. Caloric MAE is measured in calories, all others are measured in grams. Baseline is the error for an approach that always predicts the mean value for each field. }
\label{table-net-perf}
\end{table*}

\noindent\textbf{Portion Independent Model} The multi-task learning framework was first applied to predicting the caloric content and macronutrient breakdown of a dish independent of portion size. This was done by normalizing the caloric and macronutrient values by the mass of the overall dish, which converts units to calories per gram, carbohydrates per gram, fat per gram and protein per gram. In order to compare these predictions to our other models in Table~\ref{table-net-perf}, we multiply these predictions by the ground truth mass to generate caloric and macronutrient values in standard units (kilocalories and grams, respectively).

This network achieves an MAE of 9.5\% when predicting calories per gram and an average MAE of 14.7\% when predicting macronutrients per gram.

\noindent\textbf{Direct Prediction} While the portion independent model provides insight into the nutritional breakdown of a meal, a much more valuable use case is to predict the absolute values of calories and macronutrients of the dish. We apply the same architecture and multitask learning methods to directly regress these absolute values, as well as total mass. This model is able to predict calories with an MAE of 26.1\% and macronutrients with an aggregate MAE of 31.9\%.

\subsection{Depth and Volume for Portion Estimation}

\noindent\textbf{RGB-D Input} In an attempt to improve upon the direct prediction of calories and macronutrients, we also explored ways to incorporated the depth images from Nutrition5k into the training and regression process. First, we naively augmented the input image with depth as a 4th channel, which is sampled to a 3 channel tensor and input into the model. Using this RGB-D input for direct prediction method yielded a calorie MAE of 18.8\% and an aggregate macronutrient MAE ~20.9\%, as shown in Table~\ref{table-net-perf}.

\noindent\textbf{RGB with Volume Estimate Input} 
Additionally, we experimented with factoring the end-to-end nutrition understanding problem into two disjoint challenges: portion independent regression and portion estimation. We use our highly effective portion independent model from Table~\ref{table-net-perf} for the former, and train an independent mass regression network for the latter. To bolster our mass regression network, we explicitly calculate an estimate for food volume from the depth image, leveraging this as a prior input to the portion estimation model.

We follow after~\cite{ando2019depthcaloriecam} to generate food volume estimates from overhead depth images. Given the distance between camera and capture plane (35.9 cm), and the per-pixel surface area at this distance (\SI{5.957e-3} cm$^2$), we calculate the per-pixel volume measurement and sum over all food pixels (using a binary foreground/background food segmentation model) to determine the final food volume estimate. This volume estimation method is performed for each dish in the Nutrition5k collection using the overhead RealSense RGB and depth images. We then train an InceptionV3~\cite{szegedy2016rethinking} network (followed by two fully connected layers of 64 and 1 dimension) to learn mass regression directly from the input overhead RGB image, volume estimation, and per-dish food mass annotations.

\begin{table}[]
    \centering
    \begin{tabular}{l|c}
        & Total Mass MAE \\
        \hline
        Image-only & 38.1g / 29.5\% \\
        Image+volume & 29.4g / 13.7\% \\
    \end{tabular}
    \caption{Mean absolute error for mass regression from an RGB (and optionally volume estimate) input.}
    \label{tab:volume_mass}
\end{table}

In the baseline variant (Image-only), only the image is used as input during training and evaluation. We evaluate the impact of adding volume as an additional signal by concatenating the volume estimation value to the output of the InceptionV3 backbone, before the following two fully connected layers. Similar to the previous experiments, we report accuracy for mass estimation in terms of absolute and relative MAE. Despite volume being a derived estimate from the raw depth data, Table~\ref{tab:volume_mass} shows that the volume estimate prior significantly improves performance in mass estimation.

Finally, we combine the predictions from our Image+volume mass estimation with the separate portion independent per-gram model predictions to show a complete, end to end direct nutrition prediction with a calorie MAE of 16.5\% and aggregate macronutrient MAE of 26.2\%, as shown in Table~\ref{table-net-perf}.

\subsection{Analysis}
Table~\ref{table-net-perf} shows the disparity in accuracy between portion independent prediction and direct nutrition prediction. Predicting direct calories compared to calories per gram increases the MAE nearly 3x from 9.5\% to 26.1\%. This demonstrates the substantial increase in complexity of the challenge when portion estimation is required. While these portion independent models can be accurate, they still require the user to measure or weigh out their meals in order to keep an accurate nutrition diary and therefore offers limited benefits over existing online nutrition trackers such as MyFitnessPal. In our depth and volume experiments, we show how the unique large scale depth data offered in Nutrition5k can be used to approach this problem. Our first attempt of augmenting the input image with depth already shows significant improvements on our baseline 2D direct nutrition model.
However, using the depth map to calculate a volume scalar, which is provided as a separate input to direct mass regression, brings the error down from 18.7\% to 13.7\%. And combining this volume-assisted mass prediction network with our most accurate nutrition model of calories per gram gives us an end-to-end calorie prediction pipeline with an MAE or 16.5\%.
To our knowledge, no other research has produced similar accuracy in regressing nutrition values from images and depth in the generic food space. We hypothesize that this is due to the lack of a sufficiently diverse and accurately annotated dataset to facilitate such an approach prior to \textit{Nutrition5k}. 

\subsection{Human Performance}
\label{human-performance}

In many computer vision datasets, images are first collected at scale and then annotated after the fact by human raters, who are able to accomplish their annotation task given the context in the image. However, this methodology is insufficient for nutrition datasets. To evaluate the necessity of manually weighing food items at capture time, as was done in Nutrition5k, we establish a baseline for mass estimation by people visually estimating portion sizes. We follow the current behavior of many nutrition tracking app users when logging consumption, as described in Section~\ref{introduction}. For comparison, we show the resulting mass prediction of human portion estimation versus our model's predictions.

10 images were sampled from Nutri-Test, 6 with one ingredient per plate (in order: chicken, broccoli, cucumber, bell pepper, cantaloupe, pineapple) and 4 with three different ingredients per each plate (in order: oatmeal/lemon/berries, asparagus/olives/pineapple, sausage/olives/ham, and broccoli/tofu/ham). These complexity of these dishes err on the side of simplicity compared to the average dish in Nutrition5k in order to provide the human raters a better chance at estimating portions. These images were displayed to two groups: 16 amateurs (non-nutritionists) and 4 professionals (nutritionists). We asked them to estimate the mass\footnote{We additionally surveyed a third group of 32 non-nutritionists for volume estimates, but their assessments were slightly less accurate than the mass based survey.} of each ingredient present on the plate and subsequently converted these values into nutrition estimates using the same USDA \cite{food2010usda} values we used to create our dataset. Reference points were also provided to help with estimations; for example, survey instructions indicated that a US quarter and an average cell phone weigh 5.6 and 113 grams, respectively.

\begin{figure}[h]
\begin{center}
\includegraphics[width=1.0\columnwidth]{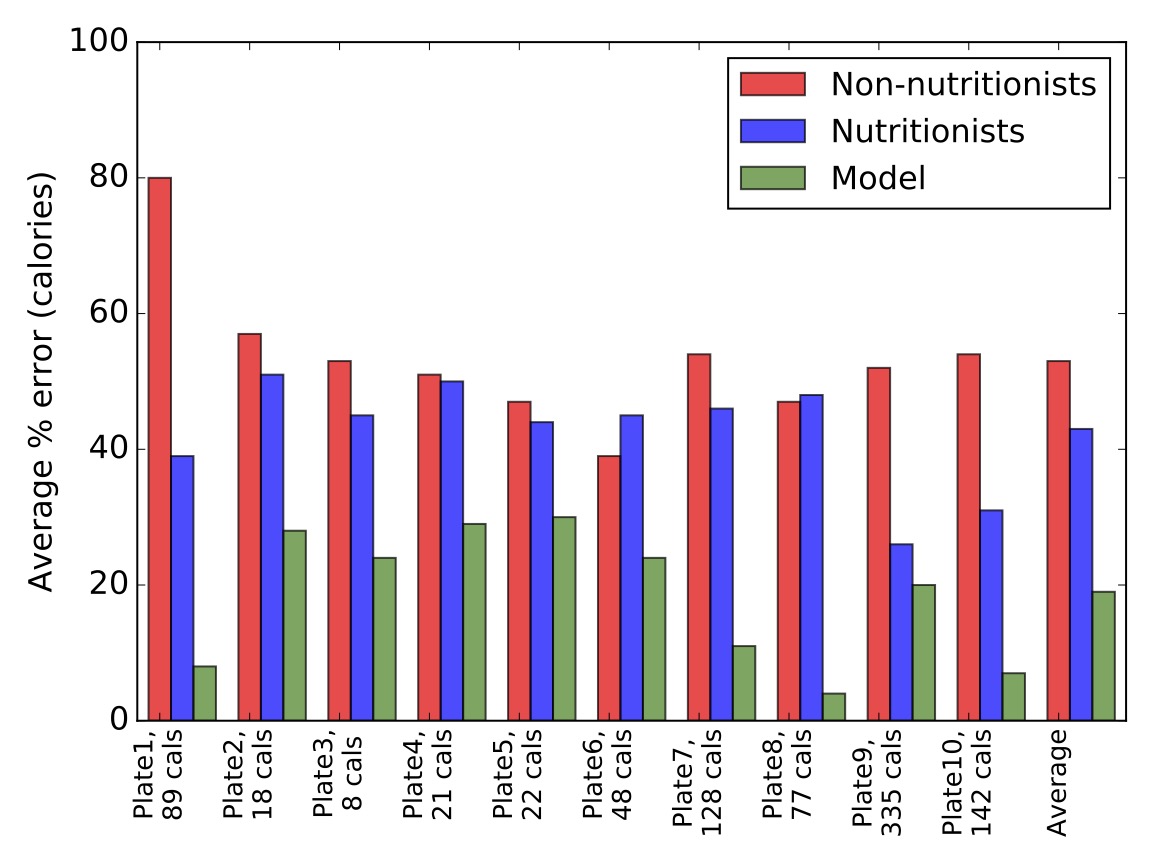}
\caption{ MAE of human estimates and MAE of model estimates as a percent of the total calories for the respective plate. The X-axis label for each plate also includes its ground truth caloric value.}
\label{human-mae-bar-plot}
\end{center}
\end{figure}

We then computed both the absolute error of each individual's total dish mass estimate utilizing known ground truth values and the mean per dish error across the two groups (non-nutritionist vs nutritionist). Figure~\ref{human-mae-bar-plot} compares mass estimation performance between human error collected from estimation surveys with the results of our 2D only direct prediction. The graph reports percent error of total mass to perform a relative comparison.

The average percent error for non-nutritionists is 53\%, which is consistent with previous reports of high human inaccuracy when performing visual portion size estimation \cite{lee2012comparison, schap2011adolescents}. The nutritionists had a lower average percent error rate of 41\%. While this survey is not an exhaustive study on the ability of human prediction performance, the magnitude of the error values this provides evidence that human rater portion annotation is not a viable method of data collection in the nutrition space.

\section{Conclusion}

In this paper, we present \textit{Nutrition5k}, the first nutritional understanding dataset of its size, diversity, and depth of labeling. We provide evidence of the challenging nature of visual annotation in our human portion size estimation studies, demonstrating the limitations of typical data annotation approaches in this space. We validate the effectiveness of our approach to data collection and the resulting \textit{Nutrition5k} dataset by training a neural network that can outperform professional nutritionists at caloric and macronutrient estimation in the \textit{generic food} setting. We further introduce multiple baseline approaches of incorporating depth data to significantly improve upon our direct nutritional prediction from 2D images alone. Our hope is that the release of this dataset will inspire further innovation in the automatic nutritional understanding space and provide a benchmark for evaluation of future techniques.

\clearpage

\renewcommand{\thesection}{\Alph{section}}

\section*{Appendix A. Additional Dataset Examples}
\begin{figure}
    \centering
    \includegraphics[clip,width=\columnwidth]{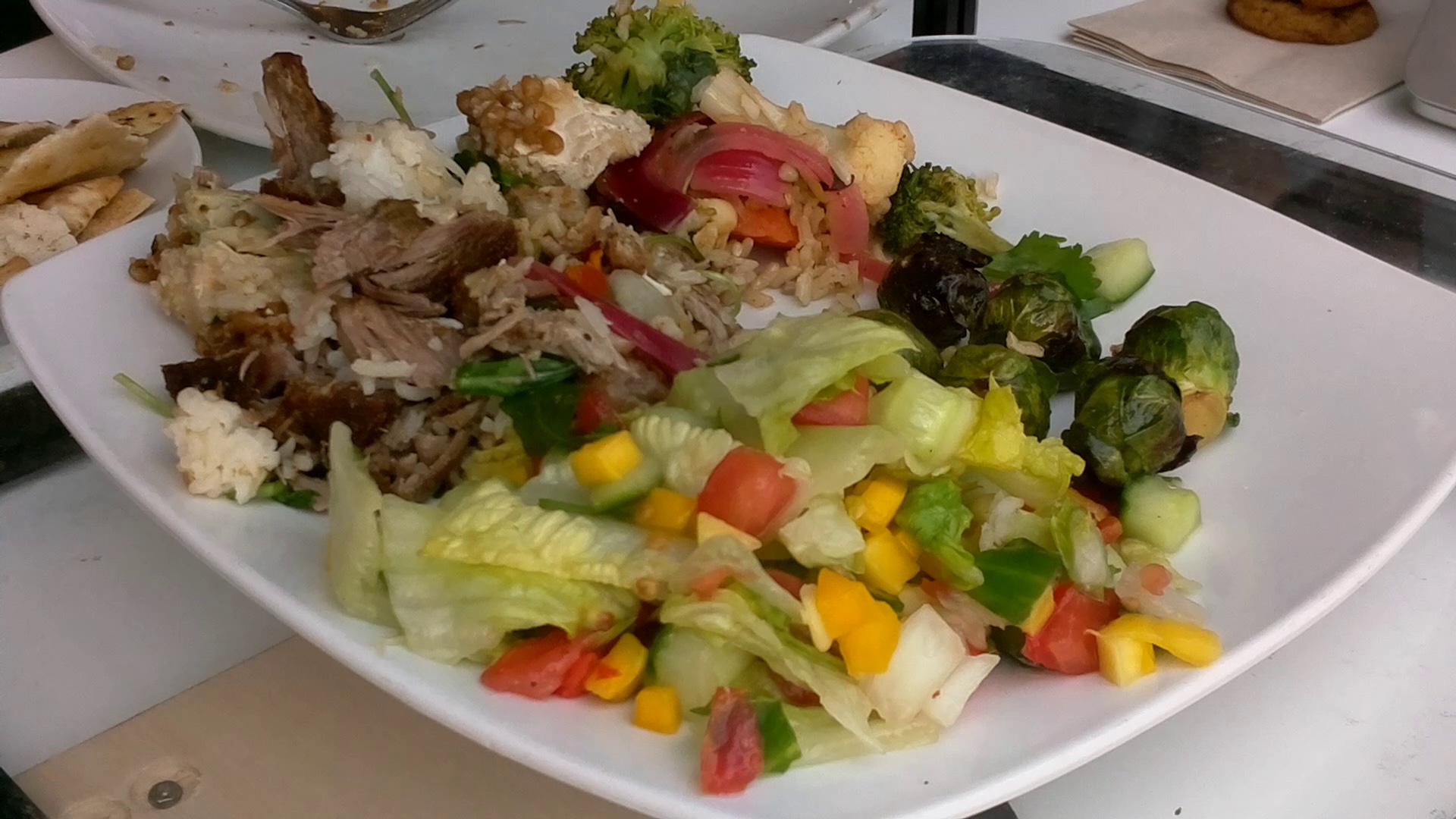}
    \caption{A single frame from one Nutrition5k dish, with fine-grained ingredient annotations (Table~\ref{tab:ingr_example}).}
    \label{fig:ingr_example}
\end{figure}

\begin{figure*}[htp]
    \centering
    \includegraphics[clip,width=2\columnwidth]{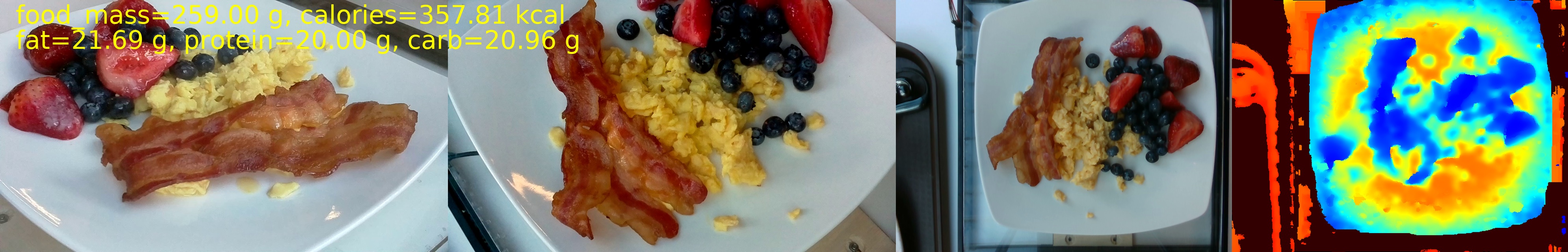}
    \includegraphics[clip,width=2\columnwidth]{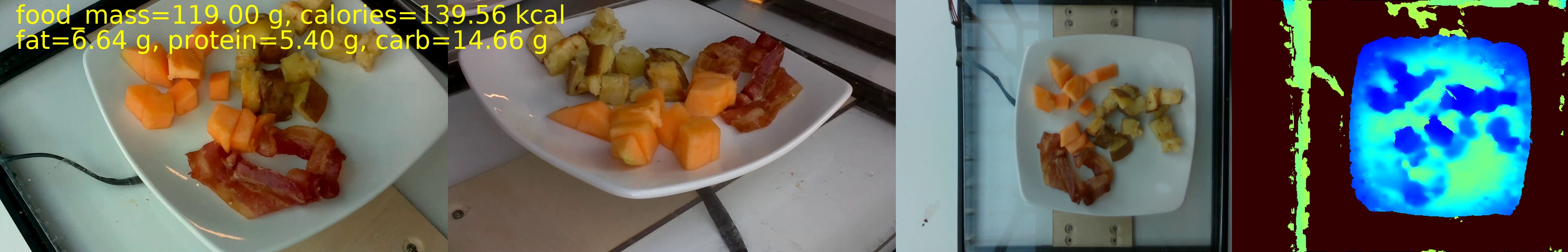}
    \includegraphics[clip,width=2\columnwidth]{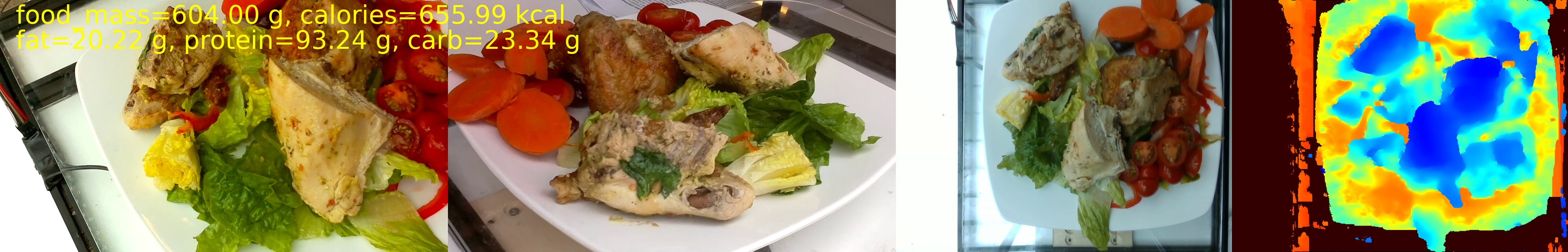}
    \includegraphics[clip,width=2\columnwidth]{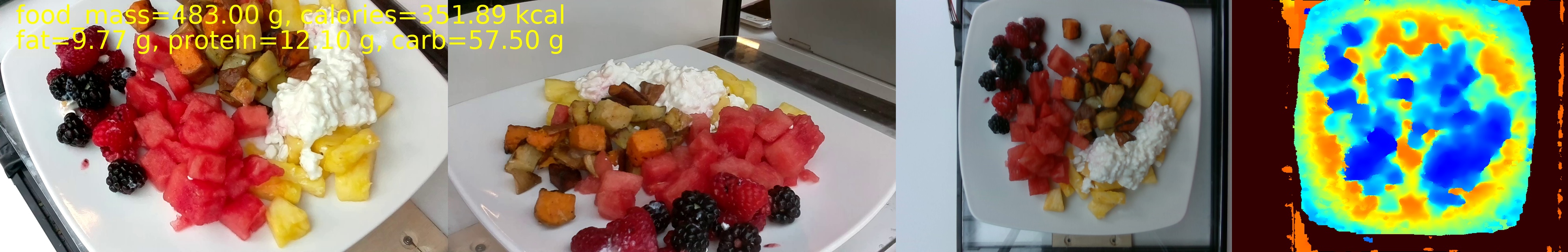}
    \includegraphics[clip,width=2\columnwidth]{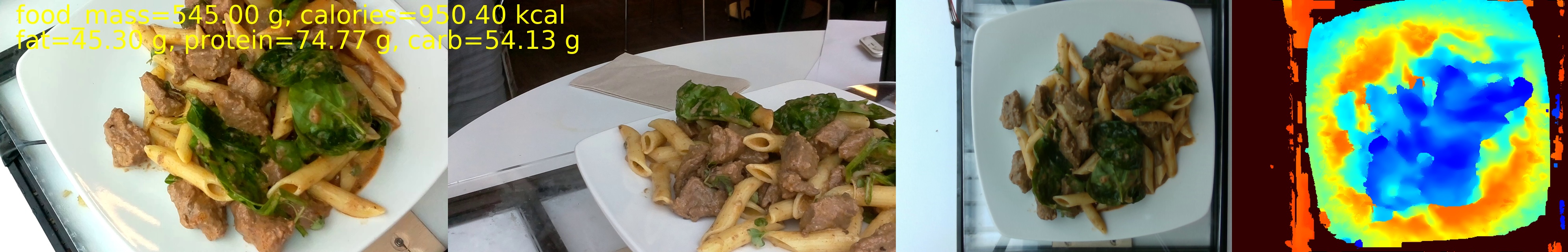}
    \includegraphics[clip,width=2\columnwidth]{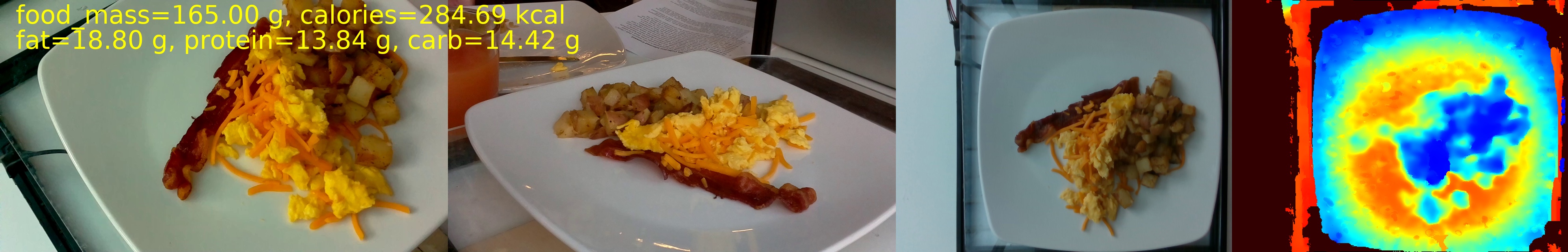}
    \includegraphics[clip,width=2\columnwidth]{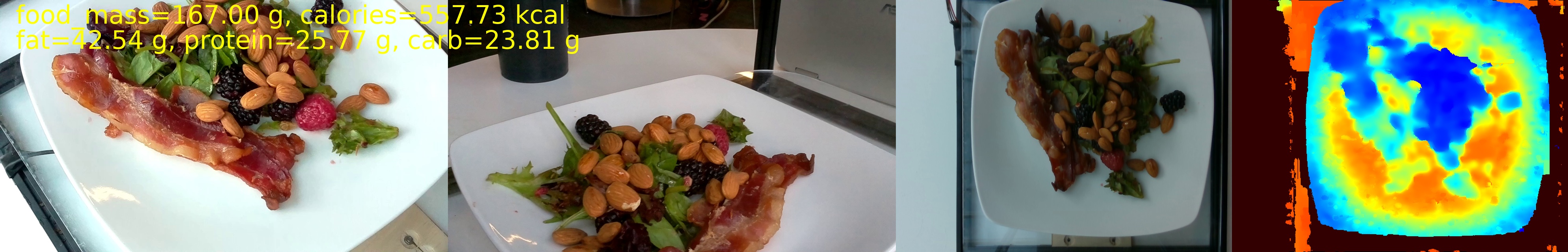}
    \caption{Some examples from the proposed \textit{Nutrition5k} dataset. Each row represents data associated with a single dish (of the roughly 5000 dishes). The first and second images are frames taken from the 30\degree and 60\degree videos, respectively. These frames represent only a single view that the rotating cameras see during their 360\degree capture of the plate. The third and fourth images are taken from the overhead Intel RealSense depth camera (depth is presented in RGB for better visualization). We also show some of the ground truth annotations associated with each dish.}
    \label{fig:n5k_examples_1}
\end{figure*}

\begin{figure*}[htp]
    \centering
    \includegraphics[clip,width=2\columnwidth]{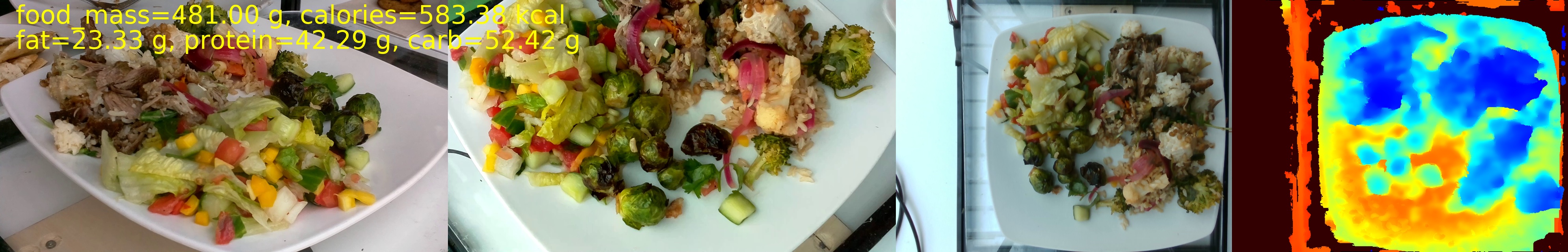}
    \includegraphics[clip,width=2\columnwidth]{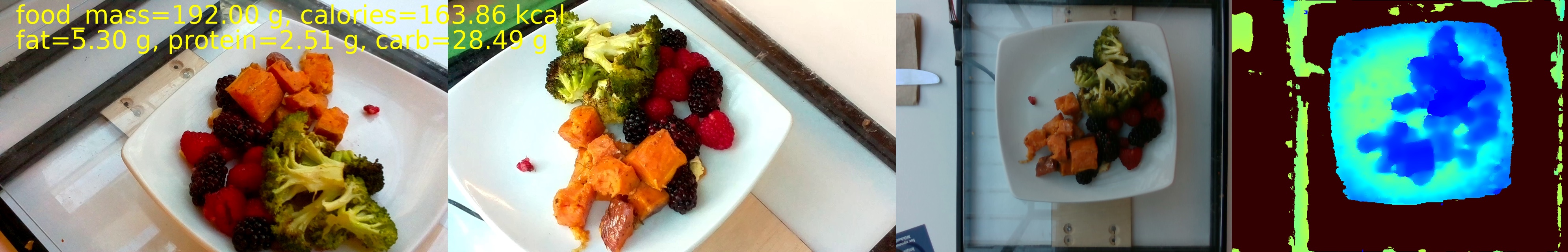}
    \includegraphics[clip,width=2\columnwidth]{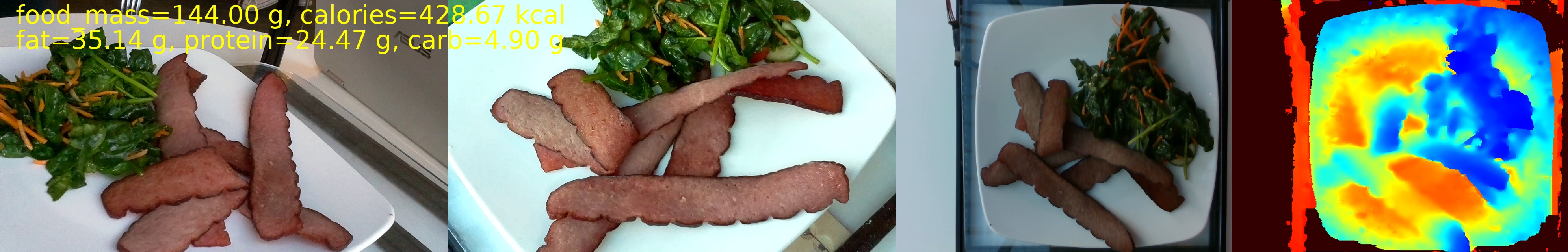}
    \includegraphics[clip,width=2\columnwidth]{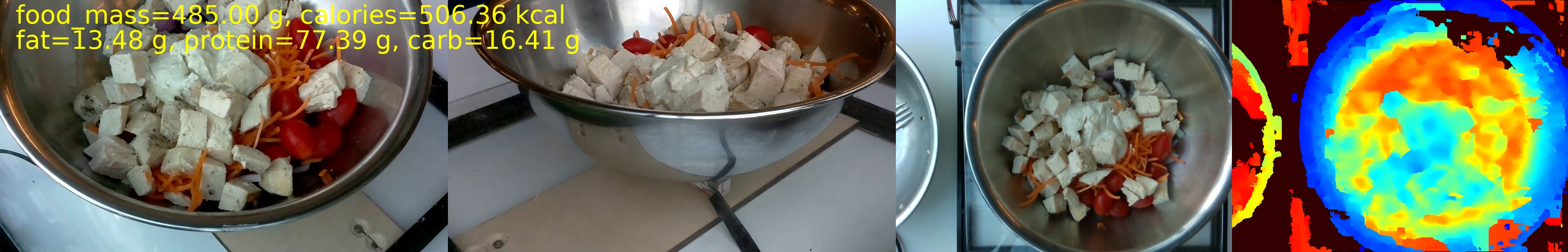}
    \includegraphics[clip,width=2\columnwidth]{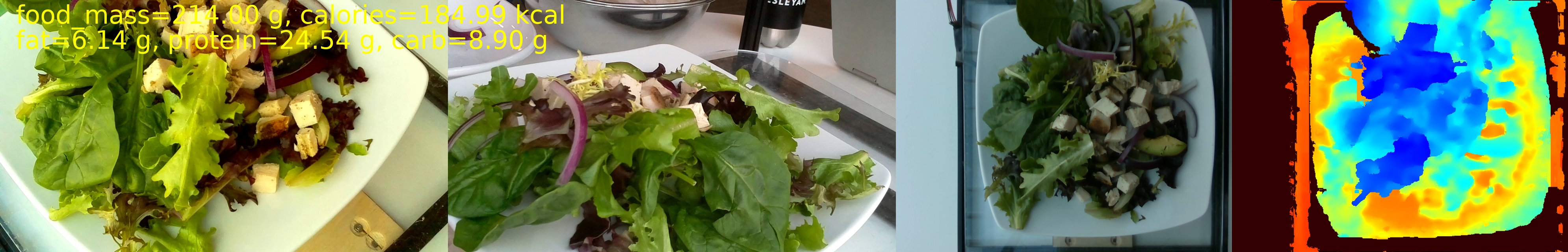}
    \caption{More examples from the proposed \textit{Nutrition5k} dataset.}
    \label{fig:n5k_examples_2}
\end{figure*}

In this supplementary section, we provide more detailed and complete exemplars from the proposed \textit{Nutrition5k} dataset. Figures~\ref{fig:n5k_examples_1} and~\ref{fig:n5k_examples_2} show frames from RGB videos, depth data, and some ground truth annotations associated with 12 unique dishes (of the roughly 5000 present in Nutrition5k). Omitted from these figures are the verbose annotations of individual ingredients present in each dish along with quantities. Instead, we provide Figure~\ref{fig:ingr_example} and Table~\ref{tab:ingr_example} to show the granularity of these ingredient annotations. We hope that these additional examples will help to convey the high quality and realistic nature of our data, and provide visualizations of the challenges presented in Section 3.6 of our paper submission.

\begin{table}[]
    \centering
    \begin{tabular}{c|c}
        \textbf{Ingredient} & \textbf{Mass} \\
        \hline
        Pork & 121.1g \\
        Fried rice & 109.0g \\
        Brussels sprouts & 66.9g \\
        Arugula & 48.7g \\
        Mustard greens & 29.8g \\
        White rice & 16.1g \\
        Mangoes & 14.9g \\
        Cucumbers & 14.9g \\
        Tomatoes & 14.9g \\
        Cherry tomatoes & 14.9g \\
        Onions & 8.6g \\
        Zucchini & 8.1g \\
        Olive oil & 3.0g \\
        Lemon juice & 2.2g \\
        Lime & 2.2g \\
        Jalapenos & 1.5g \\
        Cilantro & 1.5g \\
        Rosemary & 0.8g \\
        Garlic & 0.8g \\
        Salt & 0.7g \\
        Parsley & 0.4g \\
        Pepper & 0.1g \\
        
    \end{tabular}
    \caption{Ingredient annotations associated with Figure~\ref{fig:ingr_example}.}
    \label{tab:ingr_example}
\end{table}

\section*{Appendix B. Illustration Of Incremental Scanning}
\begin{figure*}
    \centering
    \includegraphics[clip,width=2\columnwidth]{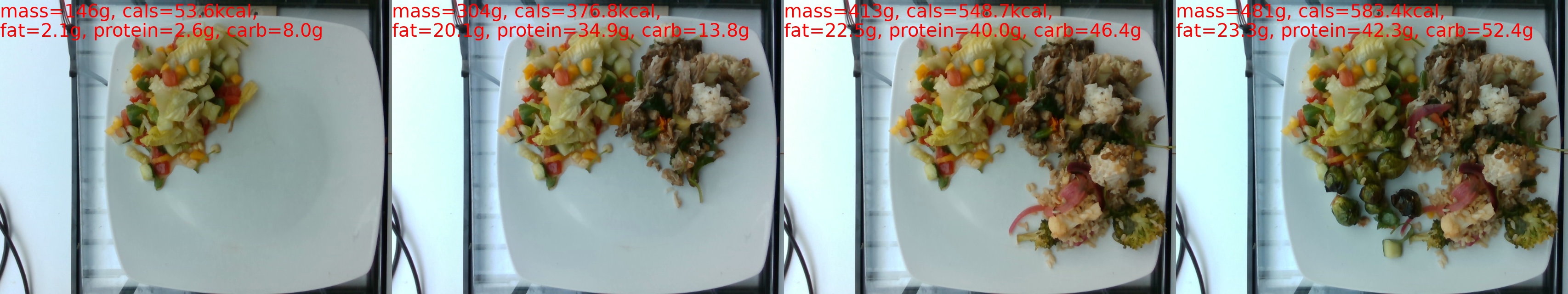}
    \caption{Illustration of the incremental scanning procedure for a single plate. Images were taken from \textit{Nutrition5k}.}
    \label{fig:incremental_scan}
\end{figure*}

Figure~\ref{fig:incremental_scan} shows an example of the incremental scanning process that we employed during data collection. One recipe (food item) is added at a time, followed by a complete device scan that captures RGB videos around the plate, depth data, and ingredient and mass annotations. Each incremental scan is represented as a unique dish within Nutrition5k, to provide more variety in our dataset. Note that our train and test sets, Nutri-Train and Nutri-Test, respect the relationship between incremental scans; all dishes belonging to the same incremental scan will exist in the same split.

\clearpage

{\small
\bibliographystyle{ieee_fullname}
\bibliography{egbib}
}

\end{document}